\documentclass[lettersize,journal]{IEEEtran}
\usepackage{amsmath,amsfonts,amssymb}
\usepackage{graphicx}
\usepackage{booktabs}
\usepackage{multirow}
\usepackage{array}
\usepackage{xcolor}
\usepackage{url}
\usepackage{cite}
\usepackage{textcomp}
\usepackage{pdflscape}
\usepackage{float}
\usepackage{dblfloatfix}
\newsavebox{\fitieeebox}
\newenvironment{fitieeetable}{\begin{lrbox}{\fitieeebox}}{\end{lrbox}%
 \ifdim\wd\fitieeebox>\textwidth
 \resizebox{\textwidth}{!}{\usebox{\fitieeebox}}%
 \else\usebox{\fitieeebox}\fi}
\begin{document}
\title{ArabGeoText: Tool-Augmented Vision--Language Models for Arabic Geospatial Text Spotting}

\author{Yakoub Bazi,
Sarah Alsaleh,
Mohamad M. Al Rahhal,
Mohamed A. Mekhtiche, and Mansour Zuair%
\thanks{Y. Bazi, S. Alsaleh, M. A. Mekhtiche, and M. Zuair are with the Computer Engineering Department, College of Computer and Information Sciences, King Saud University, Riyadh 11543, Saudi Arabia (e-mail: ybazi@ksu.edu.sa).}%
\thanks{M. M. Al Rahhal is with the Applied Computer Science Department, College of Applied Computer Science, King Saud University, Riyadh 11543, Saudi Arabia (e-mail: mmalrahhal@ksu.edu.sa).}%
\thanks{Corresponding author: Yakoub Bazi (e-mail: ybazi@ksu.edu.sa).}%
}

\markboth{}%
{Bazi \MakeLowercase{\textit{et al.}}: ArabGeoText}
\maketitle
\begin{abstract}
Geospatial text spotting involves locating and transcribing geographic labels, such as road names, district names, and landmarks, in raster maps. These labels are often small, densely arranged, and arbitrarily oriented. Although Arabic text recognition has been studied extensively in documents and natural scenes, Arabic geospatial text spotting remains underexplored. We introduce \textsc{ArabGeoText}, an Arabic-centric benchmark covering 970 geographic locations in Riyadh and Jeddah, with 17,875 manually annotated text instances per view type. For each location, we pair an OpenStreetMap (OSM) rendering with a composite created by overlaying that rendering on the corresponding remote sensing (RS) imagery. The paired views preserve label content and pixel geometry, allowing evaluation across different backgrounds. Geographic partitions support in-city and city-disjoint evaluation of detection, recognition, and end-to-end spotting. Experiments with direct Qwen3.5-4B spotting show difficulties in recovering all labels within a tile. We therefore propose a tool-augmented vision--language framework that combines YOLO26 localization and crop rectification with Qwen3.5-4B recognition adapted through low-rank adaptation (LoRA). The framework outperforms direct LoRA-adapted Qwen3.5-4B spotting across both image views and test cities. With localization held fixed, reader adaptation further improves recognition over the pretrained zero-shot reader. The dataset and code will be made publicly available at \url{https://github.com/yakoubbazi/ArabGeoText}.
\end{abstract}
\begin{IEEEkeywords}
geospatial text spotting, Arabic OCR, OpenStreetMap, vision--language models (VLMs), tool-augmented recognition
\end{IEEEkeywords}

\section{Introduction}
\label{sec:intro}

Geographic names identify roads, districts, and landmarks,
complementing the physical features visible in remote sensing
(RS) imagery. Extracting these names from raster maps and
map--imagery composites supports place-name search and geographic
image interpretation, and helps link map content to geographic
databases~\cite{kim2023mapkurator}. Geographic information systems can provide names as structured
records, but these records may be unavailable when only a rendered
image is available. In that case, names must be recovered directly
from pixels.
Geospatial text spotting addresses this task by locating
visible text instances and producing their transcriptions.

The spatial arrangement of map text differs from that of
conventional document text. Labels follow geographic features,
appear at different orientations, and compete for space
with symbols, linework, and neighboring labels. Small
characters and curved text further complicate extraction.
The ICDAR MapText competitions address related detection,
recognition, and linking tasks on historical
maps~\cite{li2024icdar, maptext2025}. In map--imagery composites,
labels also appear against the textures and structures of
the underlying RS scene. Arabic introduces
further recognition difficulties through connected letter
forms and characters distinguished by dots, which can be
difficult to resolve at small text sizes~\cite{evarest}.

Existing Arabic recognition resources cover handwritten and
printed text~\cite{khatt,apti}, natural scenes~\cite{evarest},
document understanding~\cite{kitabbench}, and synthetic
documents and calligraphy~\cite{sard2026,duwatbench2026}.
These resources cover several aspects of Arabic recognition,
but do not provide the paired map and map--imagery setting
considered here. Paired views allow
the same geographic labels to be evaluated under different
visual conditions, while geographic partitions distinguish
performance within the training city from transfer to a
city excluded from training.

We introduce \textsc{ArabGeoText}, an Arabic-centric benchmark
for geospatial text spotting covering 970 locations in
Riyadh and Jeddah, Saudi Arabia. Each location provides
paired OpenStreetMap (OSM) renderings and RS imagery
overlaid with OSM (RS+OSM), preserving identical label
content and pixel geometry. The benchmark contains
17,875 manually annotated text instances per view type,
with polygons and transcriptions for predominantly Arabic
text, together with Latin-script, mixed-script, and numeric
or symbolic labels naturally present in the maps.
The training, validation, and in-city test partitions are
drawn from Riyadh, while Jeddah is reserved for city-disjoint
testing.

We also examine the use of vision--language models (VLMs)
for geospatial text spotting.
VLMs have been applied to RS image captioning, visual
question answering, and
grounding~\cite{rsllava,earthgpt,livlmsurvey}, while text
recognition systems use visual encoders and language
decoders to recover transcriptions from image
inputs~\cite{trocr,gotocr}. Recognizing a supplied text region, however, differs from
recovering every label in a complete map tile. Direct full-tile
spotting requires the model to identify numerous small
instances, predict their boundaries, and associate each
region with a transcription in a single generated
response. Crop-level recognition allows the model to
process each localized instance separately.

We therefore propose a tool-augmented VLM framework that
separates text localization from recognition. A YOLO26 instance-segmentation
model~\cite{jocher2026yolo26} localizes text regions,
which are extracted, rectified, and resized before
recognition by Qwen3.5-4B~\cite{qwen35}. The reader is
adapted to geospatial text using low-rank adaptation
(LoRA)~\cite{hu2022lora}. We also evaluate the pretrained
reader without task-specific adaptation, using identical
detections and crop preparation to isolate the effect
of reader adaptation. For comparison, we adapt the same
VLM backbone to generate polygons and transcriptions
directly from complete tiles.

Experiments show that the proposed framework outperforms
direct LoRA-adapted Qwen3.5-4B spotting across both image
views and test cities. With localization held fixed, reader adaptation further
improves recognition over the pretrained zero-shot reader.

The main contributions are summarized as follows:
\begin{itemize}
    \item We introduce \textsc{ArabGeoText}, an Arabic-centric
    geospatial text-spotting benchmark with paired OSM and
    RS+OSM views, manually annotated polygons and transcriptions,
    and geographic partitions supporting in-city and
    city-disjoint evaluation.

    \item We propose a tool-augmented VLM framework that
    combines specialized text localization and crop
    rectification with LoRA-adapted recognition of
    geographic labels.

    \item We evaluate direct and tool-augmented spotting
    under a common protocol and analyze the effects of
    reader adaptation, joint transcription, model scale,
    and inference cost.
\end{itemize}

The remainder of this paper is organized as follows.
Section~\ref{sec:related} reviews related work.
Sections~\ref{sec:dataset} and~\ref{sec:methods} describe
the benchmark and spotting framework, respectively.
Section~\ref{sec:experiments} presents the experimental
setup and main results, Section~\ref{ssec:Discussions}
discusses the additional analyses, and
Section~\ref{sec:conclusion} concludes the paper.

\section{Related Work}
\label{sec:related}

\subsection{Map Text Spotting and Arabic Text Recognition}

Map text extraction requires the localization and recognition of labels whose orientation, spacing, and placement are influenced by geographic features. Existing systems integrate text extraction and linking within pipelines for historical maps~\cite{kim2023mapkurator}, while dedicated competitions evaluate detection, recognition, and the linking of separate words into geographic names~\cite{li2024icdar,maptext2025}. Hyper-local feature sampling and positional representations have also been used to detect and recognize long, rotated labels~\cite{palette}. Complementary approaches address text linking by combining textual content, visual features, and polygon geometry~\cite{light}. These studies highlight the roles of local appearance and spatial relationships in map understanding. Their historical-map setting, however, differs from the contemporary digital maps and map--imagery composites considered here, where the same rendered labels can be examined against different backgrounds.

Arabic text recognition has been studied across handwritten and printed text~\cite{khatt,apti}, as well as text in natural scenes~\cite{evarest}. More recent resources broaden the evaluation to document understanding~\cite{kitabbench}, synthetic book-style text~\cite{sard2026}, and calligraphic text and multimodal understanding~\cite{duwatbench2026}. These resources cover diverse Arabic recognition settings, but do not examine geographic labels in paired map and map--imagery views. In our dataset, fixed label content and geometry allow background effects to be examined, while the city-based partitions support evaluation of geographic generalization.

\subsection{Vision--Language Models for Text and Geospatial Understanding}

Vision--language approaches to text processing combine
recognition, localization, and reasoning in different ways. Encoder--decoder
models formulate text recognition as sequence generation using
pretrained visual and language representations~\cite{trocr}.
More recent approaches unify optical character recognition across
scene and document images, supporting whole-page processing and
region-level recognition guided by coordinates or visual
cues~\cite{gotocr}. Compact vision--language models also incorporate
text spotting alongside document parsing~\cite{paddleocrvl2026}.
Evaluation spans diverse text-processing tasks~\cite{ccocr}, as
well as visual document grounding and referring~\cite{dogr}.
The requirements differ across these settings. Recognition
from a supplied crop focuses on transcription, whereas full-image
spotting also requires localization and the association of each
transcription with its spatial extent. Tool-assisted
approaches obtain fine-grained evidence by invoking expert
agents to ground multimodal reasoning in visual and textual
elements~\cite{p2g}.

In RS, vision--language models have been developed
for image captioning and question answering~\cite{rsllava},
as well as multisensor interpretation~\cite{earthgpt}.
Geographic data also provide supervision for
constructing image--language training
data~\cite{lhrsbot}, training foundation models with structured
map information~\cite{geolink}, and aligning image-patch
representations with geographic semantics~\cite{osmclip}.
Other work uses paired aerial images and rendered maps to
generate geographically informed captions for domain
adaptation~\cite{osmda}. These approaches primarily use
geographic information to support image interpretation.
Related visual-grounding methods address spatial localization
through adaptive modality guidance~\cite{amvg}, unified prediction
of horizontal boxes, oriented boxes, and masks~\cite{geoground},
or progressive visual search in large scenes~\cite{geovis}.
These methods locate objects specified by referring
expressions, whereas text spotting detects and transcribes
the text visible in an image.
Our work addresses Arabic text spotting in geospatial images
by combining text-region localization with vision--language
recognition. We compare this approach with direct prediction
of text regions and transcriptions from the full image.

\section{The ArabGeoText Dataset}
\label{sec:dataset}

\textsc{ArabGeoText} covers $970$ geographic locations in
Riyadh and Jeddah, Saudi Arabia. Each location is represented by
an OSM rendering and a corresponding RS+OSM composite,
both of size $1024\times1024$ pixels. The dataset therefore
contains $1\,940$ annotated images, with $17\,875$ text
instances in each view type. The paired views preserve
the same label content and pixel geometry, allowing
comparisons of text spotting across the two backgrounds.

\subsection{Data Collection and Annotation}
\label{ssec:construction}
\label{ssec:annotation}

\begin{figure*}[!t]
\centering

\includegraphics[draft=false,width=0.73\textwidth]
{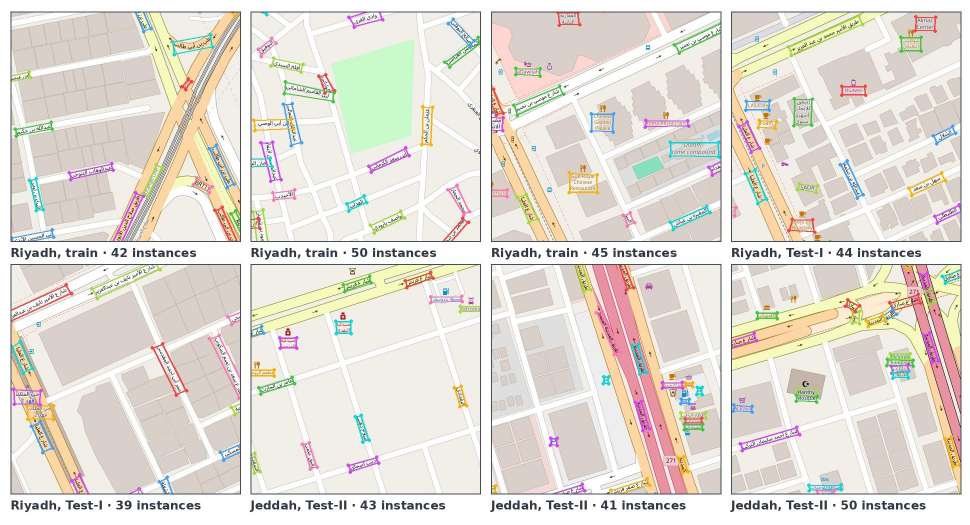}

\par\vspace{1mm}
{\small (a) OSM}

\vspace{3mm}

\includegraphics[draft=false,width=0.73\textwidth]
{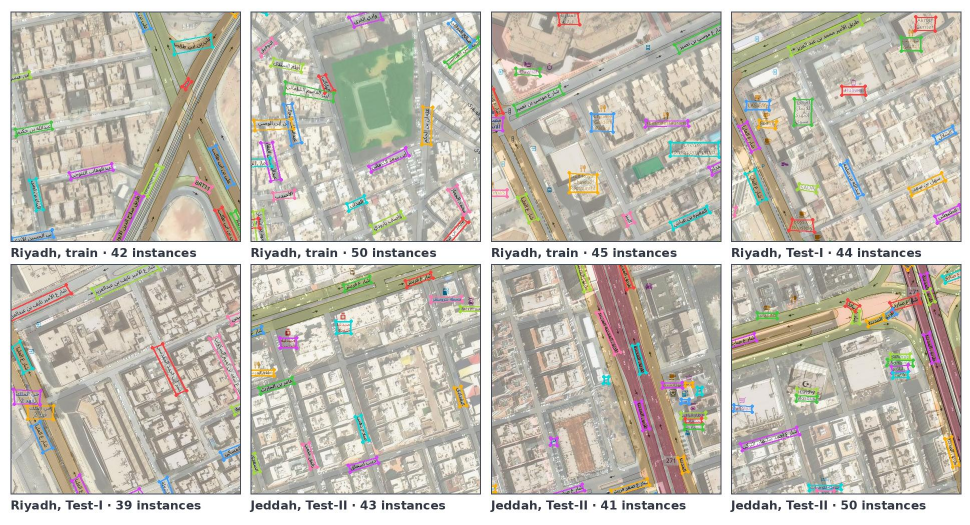}

\par\vspace{1mm}
{\small (b) RS+OSM}

\caption{Paired examples from \textsc{ArabGeoText}:
(a) OSM renderings and (b) RS+OSM composites.
Colored polygons mark annotated text instances.
The examples show cropped regions; the displayed
instance counts refer to the complete tiles.}
\label{fig:dataset_examples}
\end{figure*}

We collect standard OpenStreetMap renderings and optical
satellite imagery from Esri World Imagery over the same
geographic bounds at zoom level 18. The map renderings
contain predominantly Arabic geographic labels. We create
each RS+OSM composite by overlaying the map rendering on
the corresponding satellite image with a map opacity of
$0.63$. This value keeps geographic labels visible while
retaining details of the underlying RS image. Both views
are used for text spotting.
Figure~\ref{fig:dataset_examples} shows representative
paired examples.

Several annotators manually labeled the images using
Label Studio~\cite{labelstudio}. Each visible text instance
fully contained within a tile is annotated with a polygon
and a transcription in the displayed script. Annotators
drew polygons directly on the images, without automatic
detection proposals or annotations derived from the
underlying vector records. The polygons delineate both oriented and curved labels. The same annotation protocol was used
for both cities, and the annotation records underwent
automatic consistency checks before merging. Text clipped
by a tile boundary is not annotated; its treatment during
evaluation is described in Section~\ref{ssec:metrics}.

\label{ssec:splits}
The Riyadh locations are divided into $500$ training,
$108$ validation, and $109$ Test-I locations. All $253$
Jeddah locations are reserved for Test-II. Each location
belongs to a single partition, and its paired views remain
together. Test-I evaluates performance on held-out locations
within the training city, while Test-II evaluates transfer
to a city excluded from training.

\subsection{Statistical Analysis}
\label{ssec:stats}

Table~\ref{tab:dataset_stats} summarizes the number of images
and labels in each partition, together with label size and
orientation. The statistics are reported per view type; each location
is counted once rather than once for each paired image.
Text height is measured as the short side of the rotated
bounding rectangle.
Images from Riyadh contain an average of $20.5$--$21.7$
labels, compared with $11.6$ in Jeddah. Jeddah images
contain fewer labels, but their text is smaller.
The median height is $15.4$~px in Test-II, compared with
$18.4$~px in training. Labels below $12$~px account for
$13.4\%$ of Test-II instances, compared with $1.8\%$ in
training. Rotated labels are more common in Riyadh:
$81.5\%$--$86.2\%$ are inclined more than $15^{\circ}$
from horizontal, compared with $46.5\%$ in Jeddah.

\begin{table*}[!tbp]
\centering
\caption{Dataset statistics per view type.
Height denotes median text height; Small and Rot.\ denote
the percentages of labels below $12$~px and inclined
more than $15^{\circ}$ from horizontal, respectively.}
\label{tab:dataset_stats}
\small
\setlength{\tabcolsep}{5pt}
\begin{fitieeetable}
\begin{tabular}{llrrrrrrr}
\toprule
Partition & City & Images & Inst. &
\multicolumn{2}{c}{Inst./image} &
Height (px) & Small (\%) & Rot. (\%) \\
\cmidrule(lr){5-6}
& & & & Mean & Max & & & \\
\midrule
Train      & Riyadh & 500 & 10\,247 & 20.5 &  51 & 18.4 &  1.8 & 86.2 \\
Validation & Riyadh & 108 &  2\,326 & 21.5 & 106 & 17.9 &  3.9 & 81.5 \\
Test-I     & Riyadh & 109 &  2\,367 & 21.7 &  59 & 18.2 &  3.2 & 83.6 \\
Test-II    & Jeddah & 253 &  2\,935 & 11.6 &  50 & 15.4 & 13.4 & 46.5 \\
\midrule
Total      & ---    & 970 & 17\,875 & 18.4 & 106 & 17.9 &  4.2 & 78.7 \\
\bottomrule
\end{tabular}
\end{fitieeetable}
\end{table*}

\label{ssec:lexical}
Table~\ref{tab:lexical_stats} summarizes script composition
and vocabulary overlap with the training partition. The Unique column counts distinct transcriptions,
with repeated labels counted only once.
Arabic is the main script in all partitions. It accounts
for $91.2\%$ of training labels and $74.0\%$ of Test-II
labels. Latin-script labels are more common in Test-II,
where they account for $24.3\%$ of instances, compared
with $8.3\%$ in training.

The frequency of labels already seen in training also
differs between the two test partitions. In Test-I, $43.2\%$ of label
instances have a transcription found in training.
This proportion falls to $15.0\%$ in Test-II.
When repeated transcriptions are counted only once,
the overlap is $26.2\%$ for Test-I and $6.4\%$ for
Test-II. Most labels in Test-II therefore have
transcriptions absent from the benchmark training set.
Evaluation on Jeddah therefore involves differences
in text size, orientation, script, and vocabulary.

\begin{table*}[!tbp]
\centering
\caption{Script composition and vocabulary overlap with
training. Ar., Lat., Mix., and Num.\ denote Arabic, Latin,
mixed-script, and numeric or symbolic labels.
Overlap is reported for all instances and distinct
transcriptions.}
\label{tab:lexical_stats}
\small
\setlength{\tabcolsep}{4pt}
\begin{fitieeetable}
\begin{tabular}{lrrrrrrr}
\toprule
& \multicolumn{4}{c}{Script (\%)} & &
\multicolumn{2}{c}{Training overlap (\%)} \\
\cmidrule(lr){2-5}\cmidrule(lr){7-8}
Partition & Ar. & Lat. & Mix. & Num. &
Unique & Inst. & Unique \\
\midrule
Train      & 91.2 &  8.3 & 0.2 & 0.3 & 5\,624 & --   & --  \\
Validation & 85.9 &  9.9 & 0.1 & 4.1 & 1\,530 & 45.2 & 29.4 \\
Test-I     & 91.1 &  7.3 & 0.3 & 1.3 & 1\,566 & 43.2 & 26.2 \\
Test-II    & 74.0 & 24.3 & 1.3 & 0.4 &   981 & 15.0 &  6.4 \\
\bottomrule
\end{tabular}
\end{fitieeetable}
\end{table*}

\begin{figure*}[!t]
\centering
\includegraphics[width=0.85\textwidth]{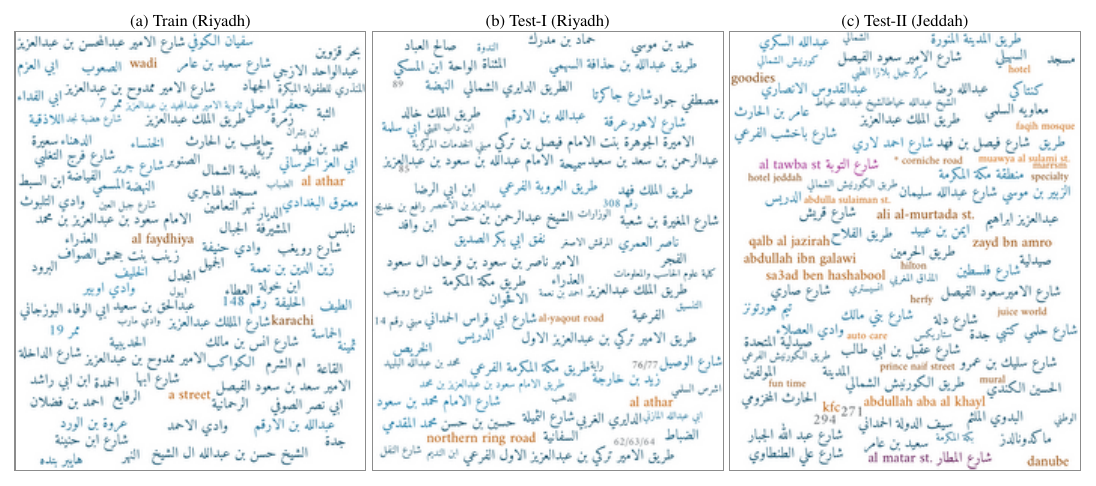}
\caption{Sample label vocabulary from training, Test-I,
and Test-II. Colors indicate Arabic (blue), Latin
(orange), mixed-script (magenta), and numeric (gray)
labels. Font size does not represent frequency.}
\label{fig:wordcloud}
\end{figure*}

Figure~\ref{fig:wordcloud} illustrates sample labels
from the training, Test-I, and Test-II partitions.
Labels are sampled across the frequency range;
font size does not indicate how often a label occurs.

\section{VLM-Based Geospatial Text Spotting}
\label{sec:methods}

We study two ways to recover geospatial text from a complete
image tile: jointly predicting locations and transcriptions
with a VLM, or recognizing regions supplied by a specialized
localization tool. We evaluate three configurations: direct VLM spotting, tool-augmented spotting
with zero-shot recognition, and tool-augmented spotting
with adapted recognition. 

\subsection{Task Formulation}
\label{ssec:task}

Let $I_i$ be an input tile in either the OSM or RS+OSM
view. Its annotation is a variable-size set
\begin{equation}
\mathcal{Y}_i
=\left\{(p_{ij},t_{ij})\right\}_{j=1}^{n_i},
\end{equation}
where $p_{ij}$ is the polygon enclosing the $j$th text
instance, $t_{ij}$ is its transcription, and $n_i$ is
the number of annotated instances. Although Arabic
predominates in \textsc{ArabGeoText}, the task includes
all annotated scripts and numeric or symbolic labels.

\begin{figure*}[!t]
\centering
\includegraphics[draft=false,width=0.73\textwidth]
{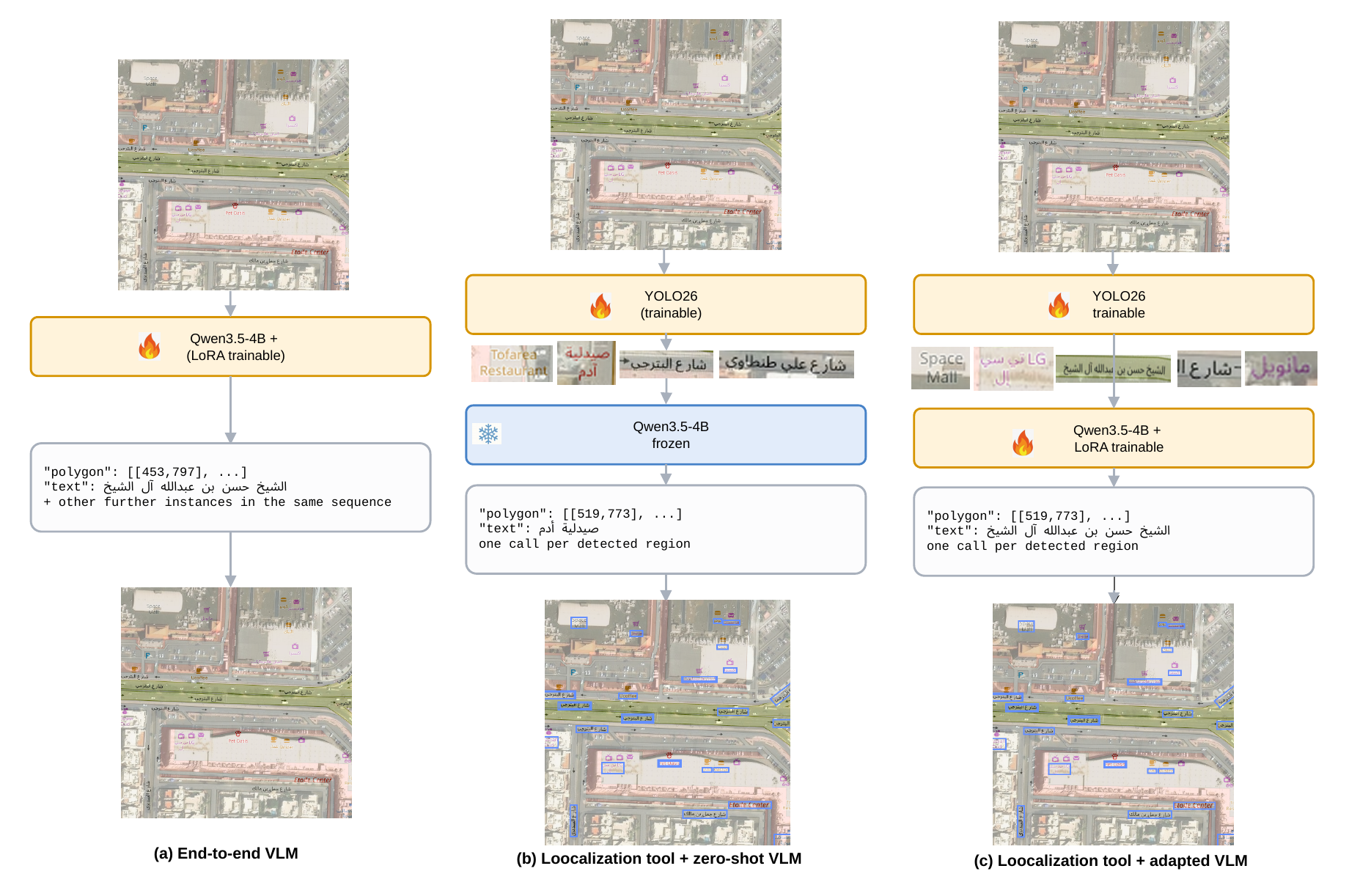}
\caption{Three paradigms for VLM-based geospatial text
spotting: (a) direct spotting, (b) tool-augmented spotting
with zero-shot recognition, and (c) tool-augmented spotting
with adapted recognition.}
\label{fig:paradigms}
\end{figure*}

Given $N_{\mathrm{tr}}$ annotated training images,
the task is to recover the locations and transcriptions
of text instances in an unseen tile.
For a test image $I$, the model predicts
\begin{equation}
\widehat{\mathcal{Y}}(I)
=\left\{(\hat{p}_j,\hat{t}_j)\right\}_{j=1}^{\hat{n}},
\end{equation}
where the predicted instance count $\hat{n}$ may differ
from the reference count. The validation partition is
used for model selection, while the test partitions
are reserved for final evaluation.
Figure~\ref{fig:paradigms} summarizes the three approaches.

\subsection{Direct VLM Spotting}
\label{ssec:e2e}

In direct spotting, the VLM generates a structured response. Qwen3.5-4B receives the full tile
and an instruction to locate and transcribe its text.
Each annotation set is serialized into a target response
$a_i$ containing the polygon coordinates and transcription
of every instance. Coordinates are expressed on a normalized
integer grid, and a common response format is used across
samples so that generated outputs can be parsed into
instance sets. Each training example consists of the
triplet $(I_i,q_{\mathrm{spot}},a_i)$, where
$q_{\mathrm{spot}}$ is the spotting instruction.

We adapt the pretrained VLM through supervised instruction
tuning with LoRA~\cite{hu2022lora}.
For a pretrained weight matrix
$\mathbf{W}_0\in\mathbb{R}^{d\times k}$, LoRA introduces
trainable matrices
$\mathbf{A}\in\mathbb{R}^{r\times k}$ and
$\mathbf{B}\in\mathbb{R}^{d\times r}$, with
$r\ll\min(d,k)$. The resulting transformation is
\begin{equation}
\mathbf{h}
=\mathbf{W}_0\mathbf{x}
+\frac{\alpha}{r}\mathbf{B}\mathbf{A}\mathbf{x},
\end{equation}
where $\alpha$ controls the update scale.
The pretrained weights remain frozen, and only the
LoRA parameters are optimized. The rank, scaling
parameter, and target modules are specified in
Section~\ref{ssec:setup}.

Let $a_i=(a_{i,1},\ldots,a_{i,L_i})$ denote the tokenized
target response, $\Theta_0$ the frozen pretrained
parameters, and $\theta_{\mathrm{L}}$ the trainable
LoRA parameters. Training minimizes the response-token
negative log-likelihood
\begin{equation}
\mathcal{L}_{\mathrm{spot}}
=-\sum_{i=1}^{N_{\mathrm{tr}}}\sum_{k=1}^{L_i}
\log p_{\Theta_0,\theta_{\mathrm{L}}}
\left(a_{i,k}\mid I_i,q_{\mathrm{spot}},a_{i,<k}\right),
\end{equation}
where $a_{i,<k}$ denotes the preceding response tokens.
This objective includes both coordinates and transcriptions,
so the model learns localization and recognition through
a single output sequence.

At inference, the adapted VLM receives an unseen tile
with the same spotting instruction. Its generated
response is parsed into
$\widehat{\mathcal{Y}}^{\mathrm{Direct}}(I)$.
Both text locations and transcriptions are thus predicted
from the full image.

\subsection{Tool-Augmented VLM Spotting}
\label{ssec:tool}

Direct generation requires the VLM to recover the geometry
and content of many small, closely spaced labels in a
$1024\times1024$ image. In the tool-augmented approach,
a specialized tool localizes these regions, and the VLM
recognizes the resulting crops. The same localization and
preprocessing procedure is applied to every tile and shared
by both reader variants.

\subsubsection{Localization and Crop Preparation}

The localization tool uses the instance-segmentation
variant of YOLO26~\cite{jocher2026yolo26}, a convolutional
model family for object detection and segmentation.
YOLO26 combines image features at multiple scales
to generate shared mask templates. These templates are
combined using instance-specific weights to produce
a separate mask for each detected region.
We train the model on the polygon annotations of the
\textsc{ArabGeoText} training partition to segment
individual text regions. It operates on the same OSM
or RS+OSM tiles as the direct VLM and predicts text
regions without producing transcriptions. The predicted masks define each text region for polygon
localization and crop preparation. The VLM then reads
the resulting crops.

\begin{figure*}[!t]
\centering
\includegraphics[draft=false,width=0.82\linewidth]
{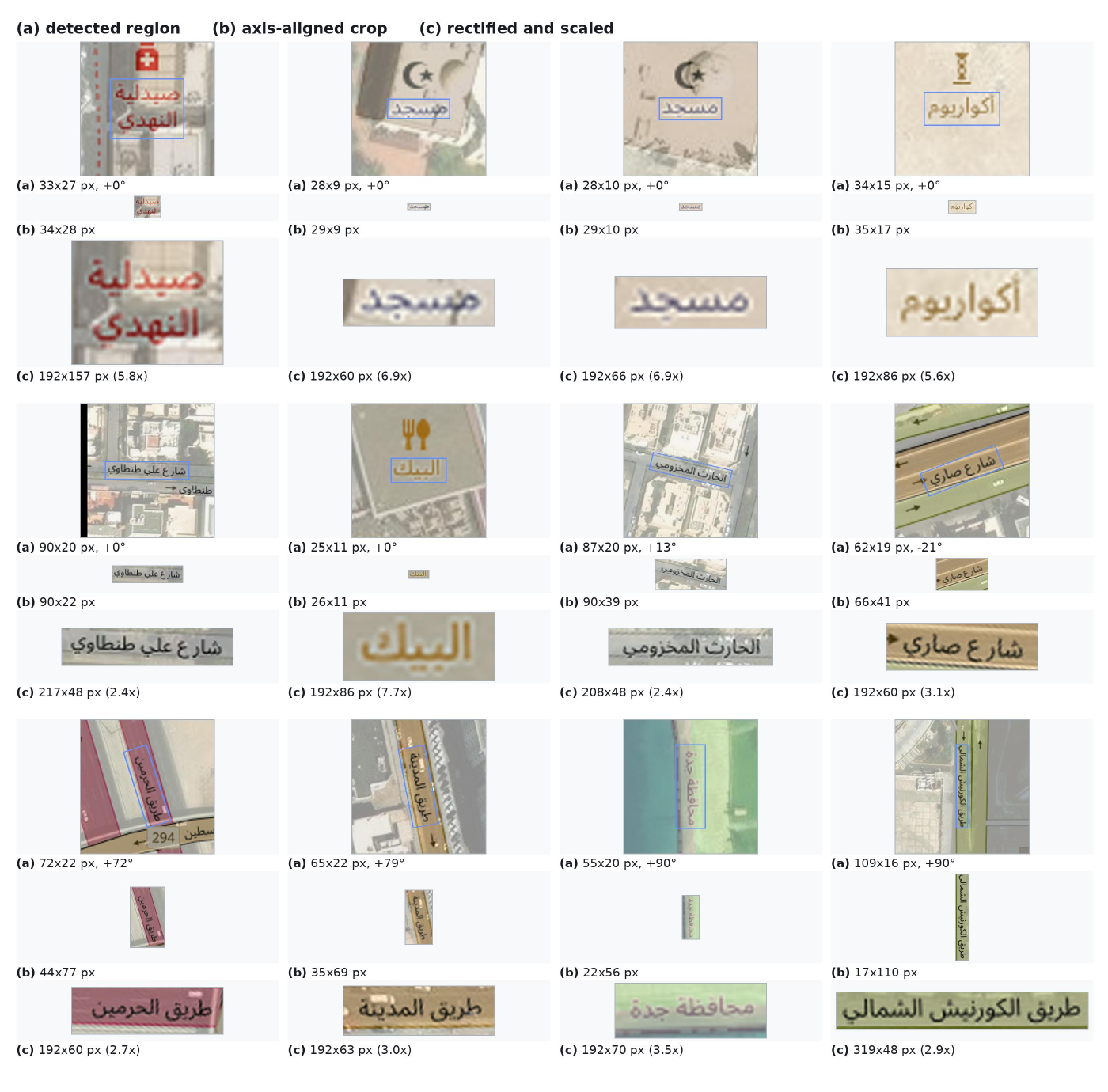}
\caption{Crop preparation: (a) text region detected by
YOLO26, (b) its axis-aligned crop, and (c) the rectified
and upscaled crop supplied to the VLM reader.}
\label{fig:crop_preprocessing}
\end{figure*}

For crop preparation, each predicted mask is enclosed
by its minimum-area oriented rectangle. The rectangle
corners are ordered consistently, and the quadrilateral
is expanded by $12\%$ of its short side before
rectification with a projective transform. The rectified
crop is then upscaled using bicubic interpolation while
preserving its aspect ratio, with a minimum height of
$48$ pixels and a minimum width of $192$ pixels.
Figure~\ref{fig:crop_preprocessing} illustrates this procedure.

We denote the tool output by
\begin{equation}
\mathcal{T}(I)
=\left\{(\hat{p}_j,c_j)\right\}_{j=1}^{\hat{n}},
\end{equation}
where $\hat{p}_j$ is the detected polygon in the original
tile and $c_j$ is its prepared recognition crop.
The oriented rectangle is used to extract and rectify
the crop; the detector polygon is retained for the
final spotting output.

\subsubsection{Zero-Shot and Adapted Recognition}

Each crop $c_j$ is processed independently by a
Qwen3.5-4B reader with a fixed recognition instruction
$q_{\mathrm{rec}}$. We consider two settings that
share the same localization and crop-preparation
procedure.

In the \emph{zero-shot recognition} setting, the reader
uses the pretrained parameters $\Theta_0$ defined in
Section~\ref{ssec:e2e}:
\begin{equation}
\hat{t}_j
=\mathcal{R}_{\Theta_0}(c_j,q_{\mathrm{rec}}).
\end{equation}
No task-specific adaptation is applied to the reader.
Here, zero-shot refers to recognition: the localization
tool is trained on the benchmark training partition.

In the \emph{adapted recognition} setting, the same
backbone is fine-tuned with LoRA using crops extracted
from the annotated text regions in the training images
and paired with their reference transcriptions.
At inference, crops are extracted from YOLO26 predictions.
The same crop-preparation procedure is used for annotated
and predicted regions.
The adapted reader predicts
\begin{equation}
\hat{t}_j
=\mathcal{R}_{\Theta_0,\theta_{\mathrm{L}}^{\mathrm{rec}}}
(c_j,q_{\mathrm{rec}}),
\end{equation}
where $\theta_{\mathrm{L}}^{\mathrm{rec}}$ denotes
the reader's trainable LoRA parameters and $\Theta_0$
remains frozen. Adaptation follows the formulation in Section~\ref{ssec:e2e},
using a crop as input and its transcription as the target
in place of the full tile and polygon--text response. The loss is evaluated only on transcription
tokens; the reader does not predict polygon coordinates.
The same recognition instruction is used during
training and inference.
For either reader, each transcription is paired with
its corresponding detector polygon. Using identical detections and crop preprocessing in the
two settings isolates the effect of reader adaptation.

\begin{table*}[!tbp]
\centering
\caption{Text detection on \textsc{ArabGeoText}.
Values are in \%.}
\label{tab:qwen_yolo_detection}
\small
\setlength{\tabcolsep}{4pt}
\begin{fitieeetable}
\begin{tabular}{@{}lllcccccc@{}}
\toprule
& & &
\multicolumn{3}{c}{Direct Qwen3.5-4B} &
\multicolumn{3}{c}{YOLO26} \\
\cmidrule(lr){4-6}\cmidrule(lr){7-9}
Input & Test set & City &
$P$ & $R$ & H-mean & $P$ & $R$ & H-mean \\
\midrule
OSM & Test-I & Riyadh
& 34.5 & 34.6 & 34.5
& 84.7 & 92.1 & \textbf{88.2} \\
OSM & Test-II & Jeddah
& 46.7 & 48.0 & 47.4
& 75.5 & 77.6 & \textbf{76.5} \\
\midrule
RS+OSM & Test-I & Riyadh
& 36.5 & 32.2 & 34.2
& 84.5 & 92.6 & \textbf{88.4} \\
RS+OSM & Test-II & Jeddah
& 38.5 & 37.3 & 37.9
& 76.6 & 80.6 & \textbf{78.6} \\
\bottomrule
\end{tabular}
\end{fitieeetable}
\end{table*}

\begin{table*}[!tbp]
\centering
\caption{Recognition and end-to-end spotting using
Qwen3.5-4B. Tool variants use YOLO26 localization.
$M$ denotes geometry-matched pairs; metric values
are in \%.}
\label{tab:recognition_comparison}
\small
\setlength{\tabcolsep}{3pt}
\begin{fitieeetable}
\begin{tabular}{@{}lllrccccc@{}}
\toprule
Input & Test set & Method & $M$ &
Exact $\uparrow$ & CER $\downarrow$ & WER $\downarrow$ &
1-NED $\uparrow$ & E2E H $\uparrow$ \\
\midrule
\multirow{3}{*}{OSM} & \multirow{3}{*}{Test-I}
& Direct VLM
& 818 & 37.2 & 21.6 & 36.1 & 75.2 & 12.8 \\
& & Tool (zero-shot)
& 2181 & 69.9 & 10.6 & 19.0 & 91.9 & 61.7 \\
& & Tool (adapted)
& 2181 & \textbf{84.3} & \textbf{4.6} & \textbf{10.2}
& \textbf{96.5} & \textbf{74.4} \\
\midrule
\multirow{3}{*}{OSM} & \multirow{3}{*}{Test-II}
& Direct VLM
& 1409 & 47.1 & 16.3 & 33.5 & 85.0 & 22.3 \\
& & Tool (zero-shot)
& 2277 & 67.9 & 18.9 & 25.3 & 85.7 & 51.9 \\
& & Tool (adapted)
& 2277 & \textbf{74.9} & \textbf{9.0} & \textbf{18.1}
& \textbf{93.5} & \textbf{57.3} \\
\midrule
\multirow{3}{*}{RS+OSM} & \multirow{3}{*}{Test-I}
& Direct VLM
& 763 & 35.5 & 25.7 & 41.0 & 73.4 & 12.2 \\
& & Tool (zero-shot)
& 2193 & 66.2 & 12.2 & 24.7 & 91.1 & 58.5 \\
& & Tool (adapted)
& 2193 & \textbf{82.5} & \textbf{4.6} & \textbf{10.7}
& \textbf{96.2} & \textbf{72.9} \\
\midrule
\multirow{3}{*}{RS+OSM} & \multirow{3}{*}{Test-II}
& Direct VLM
& 1095 & 38.8 & 20.8 & 38.9 & 80.7 & 14.7 \\
& & Tool (zero-shot)
& 2367 & 65.7 & 20.4 & 28.4 & 85.4 & 51.6 \\
& & Tool (adapted)
& 2367 & \textbf{72.5} & \textbf{9.9} & \textbf{20.0}
& \textbf{93.0} & \textbf{57.0} \\
\bottomrule
\end{tabular}
\end{fitieeetable}
\end{table*}

\section{Experiments}
\label{sec:experiments}

\subsection{Experimental Settings}
\label{ssec:setup}

We use the Riyadh training and validation partitions described
in Section~\ref{sec:dataset} for model development.
Test-I evaluates held-out performance in Riyadh, while
Test-II evaluates cross-city transfer to Jeddah.
All spotting configurations are evaluated on both OSM
and RS+OSM inputs.
For direct VLM spotting, we adapt Qwen3.5-4B with
LoRA~\cite{hu2022lora}, using rank $r=64$ and scaling
parameter $\alpha=128$. LoRA is applied to the language
decoder, while the vision encoder and multimodal projector
remain frozen. Training uses an effective batch size of
16, an initial learning rate of $1\times10^{-4}$, cosine
learning-rate decay, $5\%$ warmup, and weight decay of
$0.01$. The maximum sequence length is 8192 tokens.
We train for five epochs and select the checkpoint with
the lowest validation loss. Inference uses greedy decoding,
with polygon coordinates represented as integers on a
normalized $[0,1000]$ grid.

For tool-augmented spotting, we use the YOLO26
instance-segmentation model
\cite{jocher2026yolo26}.
The detector is trained on the combined OSM and RS+OSM
training images at $1024\times1024$ resolution, using
a single text class and a batch size of 4. Training runs
for up to 100 epochs, with an early-stopping patience
of 30 epochs. Predictions with confidence below $0.25$
are discarded. 

With the shared detector and crop-preparation procedure,
the zero-shot reader uses pretrained Qwen3.5-4B and a fixed
recognition instruction, without task-specific adaptation. The adapted reader is
fine-tuned on $20\,494$ rectified training crops paired
with their reference transcriptions. It uses the same
LoRA configuration as the direct model, an effective
batch size of 16, a learning rate of $1\times10^{-4}$
with cosine decay, and a maximum sequence length of
2048 tokens. We train for five epochs and select the
checkpoint with the lowest validation loss. The loss
is computed only on transcription tokens. Training and
inference use the same crop preprocessing and recognition
instruction. All experiments are conducted on a workstation
with two NVIDIA RTX A6000 GPUs, each with 48~GB of memory.

\subsection{Evaluation Metrics}
\label{ssec:metrics}

We evaluate text detection, text recognition, and
end-to-end text spotting.

\emph{Text detection.}
For each tile, we match predicted and reference polygons
one-to-one using polygon intersection over union (IoU).
Only pairs with IoU $\geq 0.5$ are eligible. The assignment first maximizes the number of matched pairs,
then uses their total IoU to resolve ties between assignments
with the same number of pairs. Matching uses geometry alone, and the resulting
pairs are shared across all three evaluations.
Predictions are retained even when their transcriptions
are empty.

Since text clipped by tile boundaries is not annotated, unmatched predictions
whose polygons touch the boundary are excluded after
matching. All matched predictions are scored, including
those touching the boundary.

Detection performance is measured by precision ($P$),
recall ($R$), and H-mean ($H$):
\begin{equation}
\begin{aligned} P &= \frac{TP}{TP+FP},\quad R = \frac{TP}{TP+FN},\\ H &= \frac{2TP}{2TP+FP+FN}.\end{aligned}
\end{equation}
where $TP$ is the number of matched pairs, $FP$ is the
number of unmatched predictions remaining after boundary
exclusion, and $FN$ is the number of unmatched reference
instances.

\emph{Text recognition.}
We evaluate recognition on geometry-matched pairs.
Before comparison, predicted and reference transcriptions
are normalized for letter case and spacing.
Let $\hat{s}_i$ and $s_i$ denote the normalized predicted
and reference transcriptions of the $i$th matched pair,
respectively, and let $M$ be the number of matched pairs.
We report exact-match accuracy, character error rate
(CER), word error rate (WER), and one minus normalized
edit distance (1-NED):
\begin{align}
\mathrm{Exact}
&=
\frac{1}{M}
\sum_{i=1}^{M}\mathbf{1}(\hat{s}_i=s_i),
\\
\mathrm{CER}
&=
\frac{\sum_{i=1}^{M}d_{\mathrm{c}}(\hat{s}_i,s_i)}
     {\sum_{i=1}^{M}|s_i|_{\mathrm{c}}},
\\
\mathrm{WER}
&=
\frac{\sum_{i=1}^{M}d_{\mathrm{w}}(\hat{s}_i,s_i)}
     {\sum_{i=1}^{M}|s_i|_{\mathrm{w}}},
\\
\mathrm{1\!-\!NED}
&=
\frac{1}{M}
\sum_{i=1}^{M}
\left(
1-\frac{d_{\mathrm{c}}(\hat{s}_i,s_i)}
{\max(|\hat{s}_i|_{\mathrm{c}},|s_i|_{\mathrm{c}},1)}
\right),
\end{align}
where $\mathbf{1}(\cdot)$ is the indicator function.
The distances $d_{\mathrm{c}}$ and $d_{\mathrm{w}}$
are the character-level and word-level Levenshtein
distances, and $|\cdot|_{\mathrm{c}}$ and
$|\cdot|_{\mathrm{w}}$ denote the corresponding
sequence lengths.
CER and WER are computed from the total edit errors and
reference lengths, whereas 1-NED gives equal weight
to each matched instance. Higher Exact and 1-NED and
lower CER and WER indicate better recognition.
Because recognition is scored only on matched pairs,
methods with different detections may be evaluated
on different subsets of text instances.

\subsection{Results}
\label{ssec:main_results}

Table~\ref{tab:qwen_yolo_detection} compares the detection
performance of the direct VLM and YOLO26. YOLO26 achieves
higher precision and recall in all four settings. On Test-I,
its H-mean reaches $88.2\%$ on OSM and $88.4\%$ on RS+OSM,
compared with $34.5\%$ and $34.2\%$ for the direct VLM.
The largest recall difference occurs on Test-I RS+OSM,
where YOLO26 detects $92.6\%$ of the reference instances,
compared with $32.2\%$ for the direct VLM.

\begin{figure}[!t]
\centering
\includegraphics[width=\columnwidth]
{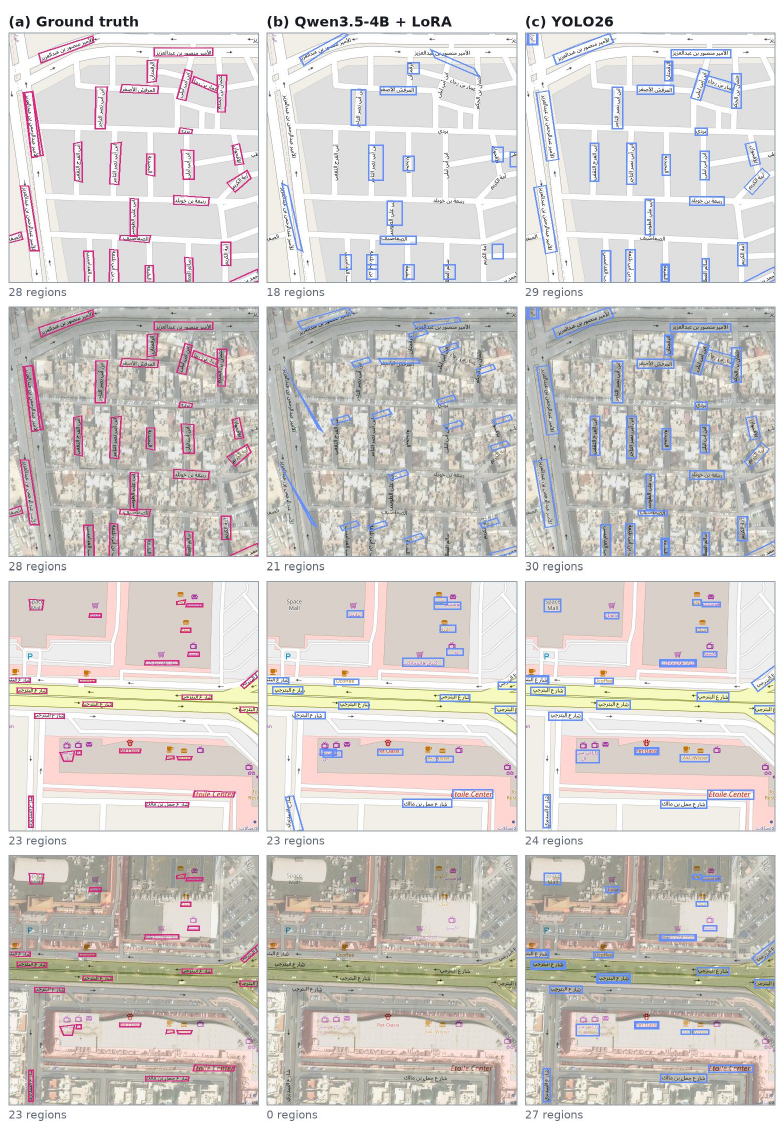}
\caption{Localization examples. Rows: Test-I OSM, Test-I
RS+OSM, Test-II OSM, and Test-II RS+OSM. Columns:
(a) ground truth, (b) adapted direct Qwen3.5-4B,
and (c) YOLO26. Reference polygons are magenta;
predictions are blue.}
\label{fig:qualitative}
\end{figure}

On Test-II, YOLO26 retains its advantage over the direct
VLM, although its H-mean falls to $76.5\%$ on OSM and
$78.6\%$ on RS+OSM. Its detection performance is similar
across the two Test-I views, with a difference of
$0.2$ percentage points. On Test-II, RS+OSM gives a
$2.1$-point higher H-mean than OSM.
Figure~\ref{fig:qualitative} compares reference polygons
and predicted text regions for both test partitions
and input views.

\begin{table*}[!tbp]
\centering
\caption{Detection with and without joint transcription.
Values are in \%.}
\label{tab:ablation_detection_only}
\small
\setlength{\tabcolsep}{3pt}
\renewcommand{\arraystretch}{1.05}
\begin{fitieeetable}
\begin{tabular}{@{}llccccccccc@{}}
\toprule
& &
\multicolumn{3}{c}{Direct VLM} &
\multicolumn{3}{c}{Detection-only VLM} &
\multicolumn{3}{c}{YOLO26} \\
\cmidrule(lr){3-5}\cmidrule(lr){6-8}\cmidrule(lr){9-11}
Input & Test set &
$P$ & $R$ & H-mean &
$P$ & $R$ & H-mean &
$P$ & $R$ & H-mean \\
\midrule
OSM & Test-I
& 34.5 & 34.6 & 34.5
& 42.0 & 48.6 & 45.1
& 84.7 & 92.1 & \textbf{88.2} \\
OSM & Test-II
& 46.7 & 48.0 & 47.4
& 47.5 & 53.5 & 50.3
& 75.5 & 77.6 & \textbf{76.5} \\
\midrule
RS+OSM & Test-I
& 36.5 & 32.2 & 34.2
& 50.5 & 49.8 & 50.1
& 84.5 & 92.6 & \textbf{88.4} \\
RS+OSM & Test-II
& 38.5 & 37.3 & 37.9
& 52.7 & 49.7 & 51.2
& 76.6 & 80.6 & \textbf{78.6} \\
\bottomrule
\end{tabular}
\end{fitieeetable}
\end{table*}

\begin{table*}[!tbp]
\centering
\caption{Effect of VLM size on recognition and spotting.
Tool variants share YOLO26 detections.
$M$ denotes geometry-matched pairs; metric values
are in \%.}
\label{tab:model_scale}
\small
\setlength{\tabcolsep}{2.5pt}
\renewcommand{\arraystretch}{1.05}
\begin{fitieeetable}
\begin{tabular}{@{}lllcrccccc@{}}
\toprule
Input & Test set & Method & VLM & $M$ &
Exact $\uparrow$ & CER $\downarrow$ & WER $\downarrow$ &
1-NED $\uparrow$ & E2E H $\uparrow$ \\
\midrule
\multirow{6}{*}{OSM} & \multirow{6}{*}{Test-I}
& Direct & 4B
& 818 & 37.2 & 21.6 & 36.1 & 75.2 & 12.8 \\
& & Direct & 0.8B
& 146 & 34.2 & 26.2 & 38.2 & 72.8 & 2.2 \\
& & Tool (zero-shot) & 4B
& 2181 & 69.9 & 10.6 & 19.0 & 91.9 & 61.7 \\
& & Tool (zero-shot) & 0.8B
& 2181 & 38.6 & 15.7 & 40.7 & 85.2 & 34.1 \\
& & Tool (adapted) & 4B
& 2181 & \textbf{84.3} & \textbf{4.6} & \textbf{10.2}
& \textbf{96.5} & \textbf{74.4} \\
& & Tool (adapted) & 0.8B
& 2181 & 76.6 & 5.3 & 13.2 & 94.9 & 67.6 \\
\midrule
\multirow{6}{*}{OSM} & \multirow{6}{*}{Test-II}
& Direct & 4B
& 1409 & 47.1 & 16.3 & 33.5 & 85.0 & 22.3 \\
& & Direct & 0.8B
& 238 & 31.1 & 22.9 & 44.9 & 77.6 & 2.4 \\
& & Tool (zero-shot) & 4B
& 2277 & 67.9 & 18.9 & 25.3 & 85.7 & 51.9 \\
& & Tool (zero-shot) & 0.8B
& 2277 & 44.7 & 29.0 & 54.3 & 83.0 & 34.2 \\
& & Tool (adapted) & 4B
& 2277 & \textbf{74.9} & \textbf{9.0} & \textbf{18.1}
& \textbf{93.5} & \textbf{57.3} \\
& & Tool (adapted) & 0.8B
& 2277 & 67.1 & 10.5 & 21.5 & 90.9 & 51.3 \\
\midrule
\multirow{6}{*}{RS+OSM} & \multirow{6}{*}{Test-I}
& Direct & 4B
& 763 & 35.5 & 25.7 & 41.0 & 73.4 & 12.2 \\
& & Direct & 0.8B
& 195 & 24.1 & 28.7 & 47.4 & 68.6 & 1.9 \\
& & Tool (zero-shot) & 4B
& 2193 & 66.2 & 12.2 & 24.7 & 91.1 & 58.5 \\
& & Tool (zero-shot) & 0.8B
& 2193 & 37.3 & 16.6 & 44.1 & 85.0 & 33.0 \\
& & Tool (adapted) & 4B
& 2193 & \textbf{82.5} & \textbf{4.6} & \textbf{10.7}
& \textbf{96.2} & \textbf{72.9} \\
& & Tool (adapted) & 0.8B
& 2193 & 72.7 & 6.2 & 15.3 & 94.1 & 64.2 \\
\midrule
\multirow{6}{*}{RS+OSM} & \multirow{6}{*}{Test-II}
& Direct & 4B
& 1095 & 38.8 & 20.8 & 38.9 & 80.7 & 14.7 \\
& & Direct & 0.8B
& 285 & 22.5 & 29.4 & 53.0 & 72.7 & 2.2 \\
& & Tool (zero-shot) & 4B
& 2367 & 65.7 & 20.4 & 28.4 & 85.4 & 51.6 \\
& & Tool (zero-shot) & 0.8B
& 2367 & 45.5 & 29.7 & 54.9 & 82.6 & 35.8 \\
& & Tool (adapted) & 4B
& 2367 & \textbf{72.5} & \textbf{9.9} & \textbf{20.0}
& \textbf{93.0} & \textbf{57.0} \\
& & Tool (adapted) & 0.8B
& 2367 & 63.2 & 11.3 & 23.6 & 90.0 & 49.6 \\
\bottomrule
\end{tabular}
\end{fitieeetable}
\end{table*}

Table~\ref{tab:recognition_comparison} reports recognition
and end-to-end spotting performance. With zero-shot
recognition, the tool-augmented pipeline achieves
end-to-end H-means of $61.7\%$ on Test-I OSM and $58.5\%$
on Test-I RS+OSM, compared with $12.8\%$ and $12.2\%$
for the direct VLM. It also achieves higher end-to-end
H-means on both Test-II views.

Recognition scores are computed on matched text instances.
On Test-I OSM, the direct VLM matches $818$ instances,
compared with $2181$ for the
tool-augmented variants. Their recognition scores therefore describe different
subsets of labels. End-to-end H-mean accounts for both
localization and transcription errors. The two
tool-augmented variants use the same detected regions,
allowing a direct comparison of zero-shot and adapted
recognition.

Reader adaptation improves all four recognition metrics
in every setting. On Test-I OSM, exact-match accuracy
increases from $69.9\%$ to $84.3\%$, while CER decreases
from $10.6\%$ to $4.6\%$. End-to-end H-mean increases
from $61.7\%$ to $74.4\%$ on OSM and from $58.5\%$
to $72.9\%$ on RS+OSM. The end-to-end gains are smaller
on Test-II: H-mean increases from $51.9\%$ to $57.3\%$
on OSM and from $51.6\%$ to $57.0\%$ on RS+OSM.
Reader adaptation also benefits the unseen city, although
performance remains below that on Test-I.

\begin{table}[!t]
\centering
\caption{Average inference time in ms over the test sets on
one NVIDIA RTX A6000.}
\label{tab:inference_efficiency}
\footnotesize
\setlength{\tabcolsep}{2pt}
\renewcommand{\arraystretch}{1.05}

\begin{tabular}{@{}lcc@{}}
\toprule
Stage & Qwen3.5-4B & Qwen3.5-0.8B \\
\midrule
YOLO26 localization    & 40.5   & 40.5 \\
Crop rectification     & 42.0   & 42.1 \\
VLM recognition        & 1619.0 & 917.7 \\
\midrule
Total per tile         & 1701.5 & 1000.3 \\
Recognition per text crop & 87.4   & 49.5 \\
\bottomrule
\end{tabular}

\end{table}

\subsection{Sensitivity analysis}
\label{ssec:Sensitivity_nalysis}

The tool-augmented pipeline achieves higher end-to-end
spotting performance than direct generation in both
test cities and input views. We examine these results further through the effect of
joint transcription on localization, the influence of VLM
size, and the trade-off between spotting accuracy and
inference time.
To assess the effect of joint transcription, we adapt
Qwen3.5-4B with LoRA to predict text polygons without
transcriptions. The remaining training and inference
settings follow those of direct spotting.


As shown in Table~\ref{tab:ablation_detection_only},
removing transcription improves detection H-mean in
all four settings. On Test-I, H-mean increases from
$34.5\%$ to $45.1\%$ on OSM and from $34.2\%$ to
$50.1\%$ on RS+OSM. However, the detection-only VLM
still falls below YOLO26 in every setting. Joint transcription therefore appears to contribute to
the localization gap, but does not fully explain it.

We next replace Qwen3.5-4B with Qwen3.5-0.8B under the same three spotting paradigms and adaptation settings. The tool-augmented variants retain the same YOLO26 detector, detected regions, and crop preparation. Table~\ref{tab:model_scale} shows that the 0.8B model
achieves end-to-end H-means of $1.9\%$--$2.4\%$ under
direct spotting, compared with $12.2\%$--$22.3\%$ for
the 4B model. With specialized localization and adapted
recognition, the 0.8B model reaches $49.6\%$--$67.6\%$.
The tool-augmented pipeline improves spotting for both
model sizes.
Reader adaptation also narrows the performance gap
between the two model sizes. On Test-I OSM, the end-to-end H-mean
difference decreases from $27.6$ percentage points
with zero-shot recognition to $6.8$ points after adaptation.
The adapted 0.8B reader also achieves higher exact-match
accuracy than the zero-shot 4B reader
($76.6\%$ versus $69.9\%$). Thus, in this setting, the adapted smaller reader
outperforms the larger pretrained reader on the same
detected regions.

Table~\ref{tab:inference_efficiency} reports inference
time, which is dominated by text recognition.
Using the 0.8B reader reduces total time from $1.70$
to $1.00$ seconds per tile, giving a $1.7\times$
speed-up. This speed-up comes with a lower end-to-end H-mean
after adaptation than that of the 4B reader.

\section{Conclusion}
\label{sec:conclusion}
We introduced \textsc{ArabGeoText}, a benchmark for Arabic
geospatial text spotting in Riyadh and Jeddah. Each location
provides an OSM rendering and a composite formed by overlaying
the rendering on satellite imagery. The paired images preserve
the same labels and their pixel locations, allowing comparison
between the two views. The benchmark supports evaluation
within the training city and in an unseen city.

We proposed a tool-augmented framework that combines YOLO26
text localization and crop rectification with LoRA-adapted
VLM recognition. The framework outperforms direct full-image
VLM spotting on both views and test cities. Reader adaptation
further improves recognition and end-to-end spotting.
Removing transcription from direct VLM generation improves
detection, but a substantial gap to YOLO26 remains.
These results support the use of specialized localization
followed by VLM recognition for this task.

Future work will investigate adaptation to new cities with fewer labeled examples and more efficient text recognition. Extending
the benchmark to additional cities would allow a broader
evaluation of geographic generalization.

\section*{Funding}
{\footnotesize
This research was funded by the Ongoing Research Funding Program, King Saud University, grant number ORF-2026-1606.\par
}

\bibliographystyle{IEEEtran}
\bibliography{refs}
\end{document}